\documentclass[10pt,journal,compsoc]{IEEEtran}
\usepackage{amsmath}
\usepackage{amsfonts}
\usepackage{mathrsfs}
\usepackage{amssymb}

\usepackage{algorithm}
\usepackage{algorithmic}

\usepackage{graphicx}

\usepackage{parskip}
\usepackage[nopar]{lipsum}

\usepackage[
            bookmarksnumbered,
            bookmarksopen,
            colorlinks,
            linkcolor=black,
            anchorcolor=black,
            citecolor=black
            ]{hyperref}

\newtheorem{thm}{Theorem}
\newtheorem{defn}{Definition}%[section]
\usepackage[compress]{cite}

\ifCLASSINFOpdf
\else
\fi
\begin{document}
\title{Fast Supervised Discrete Hashing}

\author{Jie~Gui$^*$,~\IEEEmembership{Senior Member,~IEEE,}
        Tongliang~Liu$^*$,
        Zhenan~Sun,~\IEEEmembership{Member,~IEEE,}
        Dacheng~Tao,~\IEEEmembership{Fellow,~IEEE,}
        and~Tieniu~Tan,~\IEEEmembership{Fellow,~IEEE}
\thanks{This work was supported in part by the grant of the National Science Foundation of China under Grant 61572463 and Grant 61573360, in part by the "Thirteenth Five-Year" National Key Research and Development Program of China under Grant 2016YFD0702002, in part by the grant of Strategic Priority Research Program of the Chinese Academy of Sciences under Grant XDB02080007, in part by the grant of the Open Project Program of the National Laboratory of Pattern Recognition (NLPR) under Grant 201700027, in part by the grant of the Open Project Program of the State Key Lab of CAD\&CG under Grant A1709, Zhejiang University, in part by the grant of the Shanghai Key Laboratory of Intelligent Information Processing, China under Grant IIPL-2016-003, and in part by the grant of Australian Research Council Projects under Grant FT-130101457, Grant DP-140102164, and Grant LP-150100671. All correspondence should be directed to Zhenan Sun.

J. Gui is with Institute of Intelligent Machines, Chinese Academy of Sciences, Hefei 230031, People's Republic of China. E-mail: guijie@ustc.edu.}
\thanks{T. Liu and D. Tao is with the UBTech Sydney Artificial Intelligence Institute and the School of Information Technologies in the Faculty of Engineering and Information Technologies at The University of Sydney, J12 Cleveland St, Darlington NSW 2008, Australia. E-mail: tliang.liu@gmail.com, dacheng.tao@sydney.edu.au.}
\thanks{ Z. Sun and T. Tan are with the Center for Research on Intelligent Perception and Computing, National Laboratory of Pattern Recognition, Institute of Automation, Chinese Academy of Sciences, Beijing, 100190, China and CAS Center for Excellence in Brain Science and Intelligence Technology. E-mail: \{znsun, tnt\}@nlpr.ia.ac.cn.}
\thanks{$^*$ The first two authors contribute equally to this paper.}
\thanks{ \textcopyright 20XX IEEE. Personal use of this material is permitted. Permission from IEEE must be obtained for all other uses, in any current or future media, including reprinting/republishing this material for advertising or promotional purposes, creating new collective works, for resale or redistribution to servers or lists, or reuse of any copyrighted component of this work in other works.¡±}}


\markboth{IEEE Transactions on Pattern Analysis and Machine Intelligence}%
{Shell \MakeLowercase{\textit{et al.}}: Bare Demo of IEEEtran.cls for Computer Society Journals}



\IEEEtitleabstractindextext{%
\begin{abstract}
Learning-based hashing algorithms are ``hot topics" because they can greatly increase the scale at which existing methods operate. In this paper, we propose a new learning-based hashing method called ``fast supervised discrete hashing" (FSDH) based on ``supervised discrete hashing" (SDH). Regressing the training examples (or hash code) to the corresponding class labels is widely used in ordinary least squares regression. Rather than adopting this method, FSDH uses a very simple yet effective regression of the class labels of training examples to the corresponding hash code to accelerate the algorithm. To the best of our knowledge, this strategy has not previously been used for hashing. Traditional SDH decomposes the optimization into three sub-problems, with the most critical sub-problem - discrete optimization for binary hash codes - solved using iterative discrete cyclic coordinate descent (DCC), which is time-consuming. However, FSDH has a closed-form solution and only requires a single rather than iterative hash code-solving step, which is highly efficient. Furthermore, FSDH is usually faster than SDH for solving the projection matrix for least squares regression, making FSDH generally faster than SDH. For example, our results show that FSDH is about 12-times faster than SDH when the number of hashing bits is 128 on the CIFAR-10 data base, and FSDH is about 151-times faster than FastHash when the number of hashing bits is 64 on the MNIST data-base. Our experimental results show that FSDH is not only fast, but also outperforms other comparative methods.
\end{abstract}

}

\maketitle

\IEEEdisplaynontitleabstractindextext

% \fi
\IEEEpeerreviewmaketitle

\IEEEraisesectionheading{\section{Introduction}\label{sec:introduction}}

\IEEEPARstart{T}{here} is increasing interest in large-scale visual searching in computer vision, information retrieval, and related areas due to its wide practical utility. Hashing \cite{zhang2011composite,heo2015spherical,yu2016structure,zhou2016scalable,Liu2015Multiview,Liu2016Towards,Liu2016Structure,gui2017Su,ou2013comparing} is a powerful and well-established large-scale visual search technique. Hashing generally involves generating a series of hash functions to map each example into a binary feature vector such that the produced hash codes preserve the structure of the original space (e.g., similarities between the original examples).

Existing hashing-based algorithms can be classified into two main categories: data-independent and data-dependent (learning-based). Data-independent methods do not depend on training data, instead using random projections to map examples into a feature space before binarization. Exemplars in this category include locality sensitive hashing (LSH) \cite{gionis1999similarity,korman2016coherency} and its discriminative or kernelized variants \cite{raginsky2009locality}.

In contrast, data-dependent algorithms take full advantage of training data characteristics. Various statistical learning methods have been used to map examples into binary codes for hash function learning in data-dependent hashing algorithms. Existing data-dependent hashing methods can be divided into: unsupervised, semi-supervised, and supervised methods.

In unsupervised data-dependent hashing methods, the training example labels are not required for learning. For instance, Weiss et al. \cite{weiss2009spectral} presented a spectral hashing (SH) algorithm in which the objective function was similar to Laplacian eigenmaps \cite{belkin2003laplacian}. Gong et al. \cite{gong2013iterative} proposed an iterative quantization (ITQ) algorithm that minimized the binarization loss between hash codes and the original examples. Other unsupervised data-dependent hashing methods include anchor graph hashing (AGH) \cite{liu2011hashing} and inductive manifold hashing (IMH) \cite{shen2013inductive} with t-distributed stochastic neighbor embedding (t-SNE) \cite{maaten2008visualizing}.

Semi-supervised data-dependent hashing algorithms exploit pairwise label information for hash function learning. For instance, Wang et al. \cite{wang2012semi} proposed a semi-supervised hashing (SSH) algorithm that simultaneously minimized the empirical loss for pairwise labeled training examples and maximized the variance of all training examples (both labeled and unlabeled). Kulis and Darrell \cite{kulis2009learning} proposed a binary reconstructive embedding (BRE) method that minimized the reconstruction error between the learned Hamming distance and the original Euclidean distance.

Supervised data-dependent hashing algorithms use training example labels in hash function learning. For instance, Liu et al. \cite{liu2012supervised} proposed a kernel-based supervised hashing (KSH) method that required a limited amount of label information, i.e., similar and dissimilar example pairs. Predictable dual-view hashing \cite{rastegari2013predictable} was proposed and incorporated the idea of support vector machines (SVMs) in hash learning. Other supervised learning-based hashing algorithms such as fast supervised hashing using graph cuts and decision trees (FastHash) \cite{lin2015supervised,lin2014fast} and linear discriminant analysis based hashing (LDAHash) \cite{strecha2012ldahash} have also been proposed.

Many deep learning algorithms have been proposed over the last few years, some of which have been successfully applied to many practical applications such as image classification and action recognition. Newer methods integrate deep learning and hashing \cite{lai2015simultaneous,li2015feature,liudeep} for large-scale visual searching. For example, Liong et al. \cite{erin2015deep} used deep back-propagation neural networks for hashing, while Lin et al. \cite{lin2015deep} and Zhang et al. \cite{zhang2015bit} utilized deep convolutional neural networks (CNNs) for hashing.

Hash codes are generally composed of 0 and 1 or -1 and 1. The discrete constraints imposed on the hash codes lead to mixed integer optimization problems, which are generally NP-hard. To simplify the optimization in hash learning, most hashing methods first discard discrete constraints, solve a relaxed problem, and then turn real values into the approximate hash codes by quantization (or thresholding). This relaxation strategy obviously simplifies the original binary optimization problem. However, the approximate solution is suboptimal and reduces the effectiveness of the final hash code, possibly due to the accumulated quantization error, especially when learning long hash codes. Most existing hashing algorithms fail to consider the significance of discrete optimization. In \cite{shen2015supervised}, a novel supervised discrete hashing (SDH) algorithm was proposed that directly learned the binary hash codes without relaxation. To make full use of label information, SDH was formulated as a least squares classification that regressed each hash code to its corresponding label.

The ordinary least squares regression may not, however, be optimal for classification. To further improve the performance and speed of SDH, here we propose ``fast supervised discrete hashing" (FSDH), a simple method that regresses each label to its corresponding hash code. To the best of our knowledge, this strategy has not previously been utilized in hashing. In SDH, the optimization problem is decomposed into three sub-problems, discrete optimization for hash codes being the most critical. SDH uses discrete cyclic coordinate descent (DCC) iteratively to solve discrete optimization, which is time-consuming. However, FSDH has a closed-form solution for hash learning that only requires a single step instead of iteration to solve the hash code; it is, therefore, highly efficient. When solving the projection matrix for least squares regression (another sub-problem), FSDH is usually faster than SDH. Finally, when solving the projection matrix that projects the nonlinear embedding into low-dimensional space, SDH and FSDH have similar time complexity. Therefore, FSDH is generally faster than SDH. Note that there is only a term change in the objective function of FSDH. FSDH is still non-convex and hence reaches only local minima. Our experimental results show that FSDH not only accelerates SDH but also generally outperforms SDH.

The remainder of the paper is organized as follows. We describe our proposed method in Section 2. Experimental results are presented in Section 3, and we conclude in Section 4.
\section{Our proposed method}
We first introduce the background to FSDH, i.e., SDH \cite{shen2015supervised}, in Subsection 2.1. We then introduce our proposed ``fast supervised discrete hashing" (FSDH) method in Subsection 2.2. Finally, theoretical analysis of FSDH is given in Subsection 2.3.
\subsection{Supervised discrete hashing}
Given $n$ instances $X = \{ {x_i}\} _{i = 1}^n$, our aim is to learn a set of hash codes $B = \{ {b_i}\} _{i = 1}^n \in {\{  - 1,1\} ^{n \times l}}$ to preserve their similarities in the original space, where the $i$-th row vector ${b_i}$ is the $l$-bits hash codes for ${x_i}$. The labels for all training instances are $Y = \{ {y_i}\} _{i = 1}^n \in {R^{n \times c}}$, where $c$ is the number of classes and ${y_{ik}} = 1$ if ${x_i}$ comes from class $k$ and 0 otherwise. The term ${y_{ik}}$ is the $k$-th element of $y_i$.

The objective function of SDH is:
\begin{eqnarray}\label{equ:1}
\begin{array}{l}
 \mathop {\min }\limits_{B,F,W} \sum\limits_{i = 1}^n {\left\| {{y_i} - {b_i}W} \right\|_2^2}  + \lambda \left\| W \right\|_F^2 + v\sum\limits_{i = 1}^n {\left\| {{b_i} - F\left( {{x_i}} \right)} \right\|_2^2}  \\
 s.t.\;\forall i\quad {b_i} \in {\left\{ { - 1,1} \right\}^l}. \\
 \end{array}
 \end{eqnarray}
 That is,
\begin{eqnarray}\label{equ:2}
\begin{array}{l}
 \mathop {\min }\limits_{B,F,W} \left\| {Y - BW} \right\|_F^2 + \lambda \left\| W \right\|_F^2 + v\left\| {B - F\left( X \right)} \right\|_F^2 \\
 s.t.\;B \in {\left\{ { - 1,1} \right\}^{n \times l}}. \\
 \end{array}
 \end{eqnarray}
where ${\left\|  \cdot  \right\|_F}$ is the Frobenius norm of a matrix. The first term of (\ref{equ:1}) is the ordinary least squares regression, which regress each hash code to its corresponding label. The term $W$ is the projection matrix for hash codes. The second term of (\ref{equ:1}) is for regularization. The term $F\left(  \cdot  \right)$ in the last term of (\ref{equ:1}) is a simple yet powerful nonlinear embedding to approximate the hash code
\begin{eqnarray}\label{equ:3}
F\left( x \right) = \phi \left( x \right)P,
\end{eqnarray}
where $\phi \left( x \right)$ is an $m$-dimensional row vector obtained by the Gaussian kernel $\phi \left( x \right) = [\exp \left( {{{{{\left\| {x - {a_1}} \right\|}^2}} \mathord{\left/
 {\vphantom {{{{\left\| {x - {a_1}} \right\|}^2}} \sigma }} \right.
 \kern-\nulldelimiterspace} \sigma }} \right), \cdots ,\exp \left( {{{{{\left\| {x - {a_m}} \right\|}^2}} \mathord{\left/
 {\vphantom {{{{\left\| {x - {a_m}} \right\|}^2}} \sigma }} \right.
 \kern-\nulldelimiterspace} \sigma }} \right)]$. The terms $\{ {a_j}\} _{i = 1}^m$ are the randomly selected $m$ anchor examples from the training instances, and $\sigma $ is the Gaussian kernel parameter. The matrix $P \in {R^{m \times l}}$ projects $\phi \left( x \right)$ onto the low-dimensional space. Similar formulations to equation (\ref{equ:3}) are widely utilized in other methods such as BRE \cite{kulis2009learning} and KSH \cite{liu2012supervised}.

The optimization of (\ref{equ:2}) has three steps: the F-step, which solves $P$; the G-step, which solves $W$; and the B-step, which solves $B$.

\textbf{F-step}  By fixing all other variables, the projection matrix $P$ is easily computed:
\begin{eqnarray}
P = {\left( {\phi {{\left( X \right)}^T}\phi \left( X \right)} \right)^{ - 1}}\phi {\left( X \right)^T}B.
\end{eqnarray}
\textbf{G-step}  If all other variables are fixed, it is easy to solve $W$:
\begin{eqnarray}
W = {\left( {{B^T}B + \lambda I} \right)^{ - 1}}{B^T}Y.
\end{eqnarray}
\textbf{B-step}  By fixing other variables, $B$ also has a closed-form solution. The details can be found in \cite{shen2015supervised}.

\subsection{Fast supervised discrete hashing}
%\IEEEpubidadjcol
To speed up and further improve SDH's performance, we propose a simple yet effective method called ``fast supervised discrete hashing" (FSDH). FSDH's objective function is defined as follows:
\begin{eqnarray}\label{equ:6}
\begin{array}{l}
 \mathop {\min }\limits_{B,F,W} \left\| {B - YW} \right\|_F^2 + \lambda \left\| W \right\|_F^2 + v\left\| {B - F\left( X \right)} \right\|_F^2 \\
 s.t.\;B \in {\left\{ { - 1,1} \right\}^{n \times l}}. \\
 \end{array}
 \end{eqnarray}
 SDH and FSDH only differ in the first term. SDH regresses $B$ to $Y$, while FSDH regresses $Y$ to $B$. In our view, regressing $Y$ to $B$ is the same as regressing $B$ to $Y$; the motivation for regressing $Y$ to $B$ is to accelerate the algorithm. Only the first term of FSDH's objective function makes the binary code of each class the same but, due to the third term, the binary code within each class will be different. The first term contributes to the between-class binary code differences while the third term contributes to the binary code differences of all examples.

 The problem formulated in (\ref{equ:6}) is a mixed binary integer program with three unknown variables. We use alternating optimization to iteratively solve the problem. Each iteration alternately updates $W$, $P$, $B$ ; thus, the optimization of FSDH also involves three steps, similar to SDH. The details are given below.

\textbf{F-step}  The F-step of FSDH is the same as that of SDH:
\begin{eqnarray}\label{equ:7}
P = {\left( {\phi {{\left( X \right)}^T}\phi \left( X \right)} \right)^{ - 1}}\phi {\left( X \right)^T}B.
\end{eqnarray}
\textbf{G-step}  If $B$ and $P$ are fixed, (\ref{equ:6}) can be rewritten as:
\begin{eqnarray}\label{equ:8}
\begin{array}{l}
 \mathop {\min }\limits_W \;tr\left( {{{\left( {YW - B} \right)}^T}\left( {YW - B} \right)} \right) + \lambda tr\left( {{W^T}W} \right) \\
  = \mathop {\min }\limits_W \;tr\left( {{W^T}\left( {{Y^T}Y + \lambda I} \right)W} \right) - 2tr\left( {{W^T}{Y^T}B} \right). \\
 \end{array}
 \end{eqnarray}
By setting the derivative of (\ref{equ:8}) with respect to $W$ to zero, $W$ can be solved with a closed-form solution:
\begin{eqnarray}\label{equ:9}
W = {\left( {{Y^T}Y + \lambda I} \right)^{ - 1}}{Y^T}B.
\end{eqnarray}
The time complexity of the G-step of FSDH is $O(n{c^2} + ncl)$, while the time complexity of the G-step of SDH is $O(n{l^2} + {l^3} + ncl)$. Generally, $l$ is much larger than $c$. Thus, the G-step of FSDH is usually faster than that of SDH.

\textbf{B-step}  When $F$ and $W$ are fixed, let us rewrite (\ref{equ:6}):
\begin{eqnarray}\label{equ:10}
\begin{array}{l}
 \mathop {\min }\limits_B \;tr\left( {{{\left( {B - YW} \right)}^T}\left( {B - YW} \right)} \right) \\
 \quad \quad \; + vtr\left( {{{\left( {B - F\left( X \right)} \right)}^T}\left( {B - F\left( X \right)} \right)} \right) \\
 s.t.\;B \in {\left\{ { - 1,1} \right\}^{n \times l}}. \\
 \end{array}
 \end{eqnarray}
 Since $tr\left( {{B^T}B} \right)$ is a constant, (\ref{equ:10}) is equivalent to
\begin{eqnarray}\label{equ:11}
\begin{array}{l}
 \mathop {\min }\limits_B \; - tr\left( {{B^T}\left( {YW + \nu F\left( X \right)} \right)} \right) \\
 s.t.\;B \in {\left\{ { - 1,1} \right\}^{n \times l}}. \\
 \end{array}
 \end{eqnarray}
 Thus, $B$ can be solved with a closed-form solution as follows:
\begin{eqnarray}\label{equ:12}
B = {\mathop{\rm sgn}} \left( {YW + \nu F\left( X \right)} \right).
\end{eqnarray}
The B-step of SDH involves discrete cyclic coordinate descent, so the hash code is learnt bit by bit. In contrast, the B-step of FSDH has only a single step to solve all bits, making it much faster than SDH (verified in Section 3). We present the algorithm for solving FSDH in Algorithm 1.
\begin{algorithm}
\caption{Fast supervised discrete hashing (FSDH)} \label{alg:FSSL1}
\begin{algorithmic}
%\REQUIRE ~~\\
%training examples $\{ {x_i},{y_i}\} _{i = 1}^n$; code length $l$; maximum iteration number $iteNum$; parameter $\lambda $
\STATE {\textbf{Inputs:} training examples $\{ {x_i},{y_i}\} _{i = 1}^n$; code length $l$; maximum iteration number $t$; parameter $\lambda $}
\STATE {\textbf{Output:} binary codes $\{ {b_i}\} _{i = 1}^n \in {\left\{ { - 1,1} \right\}^{n \times l}}$}
\STATE Randomly select $m$ examples $\{ {a_j}\} _{i = 1}^m$ from the training examples and get the $\phi \left( x \right)$ via the RBF kernel;
\STATE Initialize ${b_i}$ as a ${\{  - 1,1\} ^l}$ vector randomly;
\STATE Initialize $Y$ as $Y = \left\{ {{Y_{ij}}} \right\} \in {R^{n \times c}}$ where ${Y_{ij}} = \left\{ \begin{array}{l}
 1,\;if\;{y_i} = j \\
 0,\;otherwise \\
 \end{array} \right.$;
\STATE Use (\ref{equ:9}) to initialize $W$;
\STATE Use (\ref{equ:7}) to initialize $P$;
\REPEAT
\STATE \textbf{B-step} Use (\ref{equ:12}) to solve $B$;
\STATE \textbf{G-step}  Use (\ref{equ:9}) to solve $W$;
\STATE \textbf{F-step}  Use (\ref{equ:7}) to solve $P$;
\UNTIL{convergence}
%\ENSURE ~~\\
%binary codes $\{ {b_i}\} _{i = 1}^n \in {\left\{ { - 1,1} \right\}^{n \times l}}$
\end{algorithmic}
\end{algorithm}
\subsection{Theoretical analysis of FSDH}
In this subsection, we provide theoretical analysis of FSDH. Specifically, we discuss: (1) why we can replace $\|Y-BW\|_F^2$ with $\|B-YW\|_F^2$; and (2) why the proposed model FSDH is stable while the baseline SDH is not stable for learning the hash code $B$.

The proposed method replaces the term $\|Y-BW\|_F^2$ with $\|B-YW\|_F^2$ in SDH. We have already discussed how the new term reduces time complexity. However, we must also discuss other similarities and dissimilarities between the two terms. It can be seen that both the $\|B-YW\|_F^2$ and $\|Y-BW\|_F^2$ terms encourage the learned binary code to have the within-class and between-class properties that codes from the same class are similar and dissimilar otherwise. The term $\|B-YW\|_F^2$ achieves this because it encourages each code $b_i$ of $x_i$ to be picked up from $W$ according to the label $y_i$.
Thus, we conclude that $\|B-YW\|_F^2$ and $\|Y-BW\|_F^2$ are the same in the sense for generating the within-class and between-class properties and therefore could be replaced.

We discuss another beneficial property of the newly proposed term $\|B-YW\|_F^2$: it stabilizes the hashing coding algorithm.
% ; while we can show that FSD is not stable for learning the hash code $B$. Given $W$ and $Y$, minimizing $\|Y-WB\|_F^2$ with respect to $B$ is not stable because the least squares solution is unstable [REFERENCE In the SOC].
In a stable algorithm, the output hash codes do not change much if a training example is deleted or replaced with an independent and identically distributed (iid) one.
Let $S=(X,Y)=\{z_i=(x_i,y_i)\}_{i=1}^{n}$ be the training sample for FSDH and $S^i$ be the sample with the $i$-th example $z_i=(x_i,y_i),i=1,\ldots,n$ in $S$ replaced with an iid one $z'_i=(x'_i,y'_i)$.

\begin{defn}
A hashing coding algorithm is $\beta(n)$-stable if the following holds
\begin{eqnarray*}
\forall S,S^i, z_i,z'_i,i=1,\ldots,n, \|B(S)-B(S^i)\|_F\leq \beta(n),
\end{eqnarray*}
where $B(S)$ and $B(S^i)$ are the hash codes learned by employing $S$ and $S^i$, respectively, and $\beta(n)$ converges to zero with respect to the sample size $n$.
\end{defn}

FSDH is optimized by employing an alternating iteration method. We assume that the optimization algorithm stops with $K$ iterations and in the $k$-th iteration, where $k=1,\ldots,K$, $B^k, W^k, F^k$ are obtained. We can prove that FSDH for learning $W$ is stable in each iteration because of the $\ell_2$-regularization $\|W\|_F^2$.

Before presenting our result, we first modify the objective function in (6) to ensure that the regularization parameters are invariant to the sample size $n$, class size $c$, and code length $l$. The modified model is as follows:
\begin{eqnarray}\label{modifiedmodel}
%\begin{array}{l}
\min_{B,F,W}\frac{1}{nl}\|B-YW\|_F^2+\frac{\lambda'}{cl}\|W\|_F^2+\frac{\nu'}{nl}\|B-F(X)\|_F^2.
 %\end{array}
\end{eqnarray}
Note that the objective function in (\ref{modifiedmodel}) is identical to that in (6) by letting $\lambda=\lambda' n/c$ and $\nu=\nu'$.

\begin{thm}\label{main}
In the $k$-th iteration, given $B^{k}$ and $F^{k-1}$, FSDH defined in (\ref{modifiedmodel}) is stable when learning $W^k$. For any $S$ and $S^i$, let $W^k(S)$ and $W^k(S^i)$ be learned by employing the sample $S$ and $S^i$ in the $k$-th iteration, respectively. Assume that for any learned $B$ and $W$, we have $\|b-yW\|_2\leq M$, where $M$ is a universal constant. Then,
\begin{eqnarray}
\|W^k(S)-W^k(S^i)\|_F\leq 2cM/\lambda' n.
\end{eqnarray}
\end{thm}

We need the following Bregman matrix divergence \cite{watson1992characterization} to prove the theorem.
\begin{defn}
For any matrix $A$ and $B$ of the same size, the Bregman matrix divergence with respect to function $f$ is defined as
\begin{eqnarray*}
\text{Bgm}_f(A,B)=f(A)-f(B)-\text{tr}\left(\nabla f(B)^T(A-B)\right),
\end{eqnarray*}
where $\nabla f(B)$ denotes the derivative of $f$ at $B$.
\end{defn}

It is proven that if function $f$ is convex, the Bregman divergence will be non-negative and additive. For example, $\text{Bgm}_f(A,B)\geq 0$ and $\text{Bgm}_{f+g}(A,B)=\text{Bgm}_f(A,B)+\text{Bgm}_g(A,B)$ if $f$ and $g$ are both convex.

{\bf Proof of Theorem \ref{main}.} In the $k$-th iteration, let
\begin{align*}
&f_S(W)\\
&=\frac{1}{nl}\|B^k-YW\|_F^2+\frac{\lambda'}{cl}\|W\|_F^2+\frac{\nu'}{nl}\|B^k-F^{k-1}(X)\|_F^2
\end{align*}
and
\begin{eqnarray*}
r_S(W)=\frac{\lambda'}{cl}\|W\|_F^2.
\end{eqnarray*}
According to the non-negative and additive properties of Bregman divergence, we have
\begin{align}\label{bregman1}
&\text{Bgm}_{f_{S^i}}(W^k(S),W^k(S^i))+\text{Bgm}_{f_S}(W^k(S^i),W^k(S))\\
&\geq \text{Bgm}_{r_{S^i}}(W^k(S),W^k(S^i))+\text{Bgm}_{r_S}(W^k(S^i),W^k(S)).\nonumber
\end{align}

We also have
\begin{eqnarray}\label{bregman2}
&&\text{Bgm}_{r_{S^i}}(W^k(S),W^k(S^i))+\text{Bgm}_{r_S}(W^k(S^i),W^k(S))\nonumber\\
&&=\frac{\lambda'}{cl}\|W^{k}(S)\|_F^2-\frac{\lambda'}{cl}\|W^{k}(S^i)\|_F^2\nonumber\\
&&\ \ \ -2\frac{\lambda'}{cl}\text{tr}\left(W^k(S^i)^T(W^{k}(S)-W^{k}(S^i))\right)\nonumber\\
&&\ \ \ +\frac{\lambda'}{cl}\|W^{k}(S^i)\|_F^2-\frac{\lambda'}{cl}\|W^{k}(S)\|_F^2\nonumber\\
&&\ \ \ -2\frac{\lambda'}{cl}\text{tr}\left(W^k(S)^T(W^{k}(S^i)-W^{k}(S))\right)\nonumber\\
&&=\frac{2\lambda'}{cl}\|W^{k}(S^i)-W^{k}(S)\|_F^2
\end{eqnarray}
and that
\begin{eqnarray}\label{bregman3}
&&\text{Bgm}_{f_{S^i}}(W^k(S),W^k(S^i))+\text{Bgm}_{f_S}(W^k(S^i),W^k(S))\nonumber\\
&&=\frac{1}{nl}\|B^k-Y^iW^{k}(S)\|_F^2+\frac{\lambda'}{cl}\|W^{k}(S)\|_F^2\nonumber\\
&&\ \ \ +\frac{\nu'}{nl}\|B^k-F^{k-1}(X^i)\|_F^2-\frac{1}{nl}\|B^k-Y^iW^{k}(S^i)\|_F^2\nonumber\\
&&\ \ \ -\frac{\lambda'}{cl}\|W^{k}(S^i)\|_F^2-\frac{\nu'}{nl}\|B^k-F^{k-1}(X^i)\|_F^2\nonumber\\
&&\ \ \ +\frac{1}{nl}\|B^k-YW^{k}(S^i)\|_F^2+\frac{\lambda'}{cl}\|W^{k}(S^i)\|_F^2\nonumber\\
&&+\frac{\nu'}{nl}\|B^k-F^{k-1}(X)\|_F^2-\frac{1}{nl}\|B^k-YW^{k}(S)\|_F^2\nonumber\\
&&\ \ \ -\frac{\lambda'}{cl}\|W^{k}(S)\|_F^2-\frac{\nu'}{nl}\|B^k-F^{k-1}(X)\|_F^2\nonumber\\
&&=\frac{1}{nl}\|B^k-YW^{k}(S^i)\|_F^2-\frac{1}{nl}\|B^k-Y^iW^{k}(S^i)\|_F^2\nonumber\\
&&\ \ \ +\frac{1}{nl}\|B^k-Y^iW^{k}(S)\|_F^2-\frac{1}{nl}\|B^k-YW^{k}(S)\|_F^2\nonumber\\
&&=\frac{1}{nl}\|b_i^k-y_iW^{k}(S^i)\|_2^2-\frac{1}{nl}\|b_i^k-y'_iW^{k}(S^i)\|_2^2\nonumber\\
&&\ \ \ +\frac{1}{nl}\|b_i^k-y'_iW^{k}(S)\|_2^2-\frac{1}{nl}\|b_i^k-y_iW^{k}(S)\|_2^2\nonumber\\
&&\leq \frac{2M}{nl}\|y_i(W^{k}(S^i)-W^{k}(S))\|_2\nonumber\\
&&\ \ \ +\frac{2M}{nl}\|y'_i(W^{k}(S^i)-W^{k}(S))\|_2\nonumber\\
&&\leq \frac{4M}{nl}\max\{\|w_j^{k}(S^i)-w_j^{k}(S)\|_2|j=1,\ldots,c\}\nonumber\\
&&\leq \frac{4M}{nl} \|(W^{k}(S^i)-W^{k}(S))\|_F
\end{eqnarray}
where the first equality holds because $\nabla_W f_{S^i}(W^k(S^i))=\nabla_W f_{S}(W^k(S))=0$.

Combining (\ref{bregman1}), (\ref{bregman2}), and (\ref{bregman3}), we have
\begin{eqnarray*}
\frac{2\lambda'}{cl}\|W^{k}(S^i)-W^{k}(S)\|_F^2\leq \frac{4M}{nl} \|(W^{k}(S^i)-W^{k}(S))\|_F,
\end{eqnarray*}
which implies that
\begin{eqnarray*}
\|W^{k}(S^i)-W^{k}(S)\|_F\leq 2cM/\lambda' n.
\end{eqnarray*}

This completes the proof.\hfill $\blacksquare$

Since the newly proposed term $\|B-YW\|_F^2$ encourages each code $b_i$ of $x_i$ to be picked up from $W$ according to the label $y_i$, the learning algorithm being stable with respect to $W$ implies that it is also stable with respect to code $B$. Theorem \ref{main} then implies that the difference between $W^k(S)$ and $W^k(S^i)$, as well as the difference between and $B^k(S)$ and $B^k(S^i)$, will decrease as the sample size $n$ increases. Note that although the SDH algorithm is stable with respect to learning $W$ in each step, it is not stable for learning the hash code $B$ because least squares solution is not stable \cite{hoerl1970ridge}. Note also that Bousquet and Elisseeff \cite{bousquet2002stability} proved that stable algorithms will generalize well and that our empirical results in Section 3 support our theoretical analysis by showing that the newly proposed methods generalize well on the test samples.

\section{Experiments}
In this section, we demonstrate the effectiveness of our proposed method by conducting experiments on two large-scale image datasets (CIFAR-10\footnote{http://www.cs.toronto.edu/\~{}kriz/cifar.html}  and MNIST\footnote{http://yann.lecun.com/exdb/mnist/}) and a challenging and large-scale face dataset FRGC. Experiments are performed on a server with an Intel Xeon processor (2.80 GHz), 128GB RAM, and configured with Microsoft Windows Server 2008 and MATLAB 2014b.

We compare our proposed method with representative hashing algorithms including BRE \cite{kulis2009learning}, SSH \cite{wang2012semi}, KSH \cite{liu2012supervised}, FastHash \cite{lin2015supervised,lin2014fast}, AGH \cite{liu2011hashing}, and IMH \cite{shen2013inductive} with t-SNE \cite{maaten2008visualizing}. For iterative quantization (ITQ) \cite{gong2013iterative,gong2011iterative} both its supervised (CCA-ITQ) and unsupervised (PCA-ITQ) versions are utilized. CCA-ITQ uses canonical correlation analysis (CCA) for preprocessing. The public MATLAB codes and model parameters suggested by the corresponding authors are used. For fair comparison, in FSDH and SDH, we empirically set $\lambda$, $v$, and the maximum iteration number $t$  to 1, 1e-5, and 5, respectively, as in \cite{shen2015supervised}. For AGH, IMH, SDH, and FSDH, 1,000 randomly sampled anchor points are utilized.

We report the experimental results using Hamming ranking (mean of average precision, MAP), hash lookup (precision, recall, and F-measure of Hamming radius 2), accuracy, training time, and test time. The F-measure is defined as 2$\times$precision$\times$recall/(precision + recall). We also use the following evaluation metric to measure performance: precision at $N$ examples which is the percentage of true neighbors among the top $N$ retrieved examples. Note that a query is considered as a false instance if no example is returned when calculating precisions. The labels of the examples are defined as the ground truths.

\subsection{Experimental results on CIFAR-10}
\begin{table*}
\begin{center}
\scalebox{1}[1]{
\begin{tabular}{|l|c|c|c|c|c|c|c|}
\hline
Method & precision@$r$=2 & recall@$r$=2 & F-measure@$r$=2 & MAP & accuracy & training time & test time\\
\hline\hline
FSDH & \textbf{0.3142} & \textbf{0.0793} & \textbf{0.1266} & 0.4639 & 0.658 & 32.8 & 7.4e-6\\
SDH  & 0.3017  & 0.0675  &  0.1103 & 0.4668  & 0.649  & 406.4  &  6.8e-6 \\
BRE  & 0.0080  & 1.4e-6  & 2.7e-6  &  0.1640 &  0.454 &  3107.5 & 4.2e-5  \\
KSH  & 0.0291  & 3.8e-4  & 7.5e-4  &  0.4823 &  0.57 &  10534 &  8.4e-5 \\
SSH  & 0.1472  &  1.2e-4 & 2.5e-4  &  0.2222 &  0.463 & 132.7  & 1.1e-5  \\
CCA-ITQ  & 0.1899  & 0.0029  &  0.0056 & 0.3410  & 0.568  & 35.9  & 3.1e-7  \\
FastHash  & 0.1010  & 0.0138  & 0.0243  & \textbf{0.6802}  & \textbf{0.683}  & 1183.3  &  3.8e-4 \\
PCA-ITQ  & 1.0e-3  & 1.7e-7  &  3.4e-7 & 0.1803  & 0.483  & 20.0  &  \textbf{2.7e-7} \\
AGH  &  0.2502 &  3.3e-4 &  6.6e-4 & 0.1478  & 0.441  & \textbf{8.4}  &  1.1e-4 \\
IMH  &  0.1599 &  0.0021 &  0.0042 &  0.1811 &  0.365 &  59.8 & 5.3e-5  \\
\hline
\end{tabular}}
\end{center}
\caption{Experimental results on the CIFAR-10 database when the number of hashing bits is 128. Training and test times are in seconds. The best results are highlighted in bold face.}\label{tab:1}
\end{table*}

\begin{figure}
\begin{center}
%%\fbox{\rule{0pt}{2in} \rule{.9\linewidth}{0pt}}
%\includegraphics[scale=0.4]{Fig1.pdf}
\scalebox{0.36}{\includegraphics{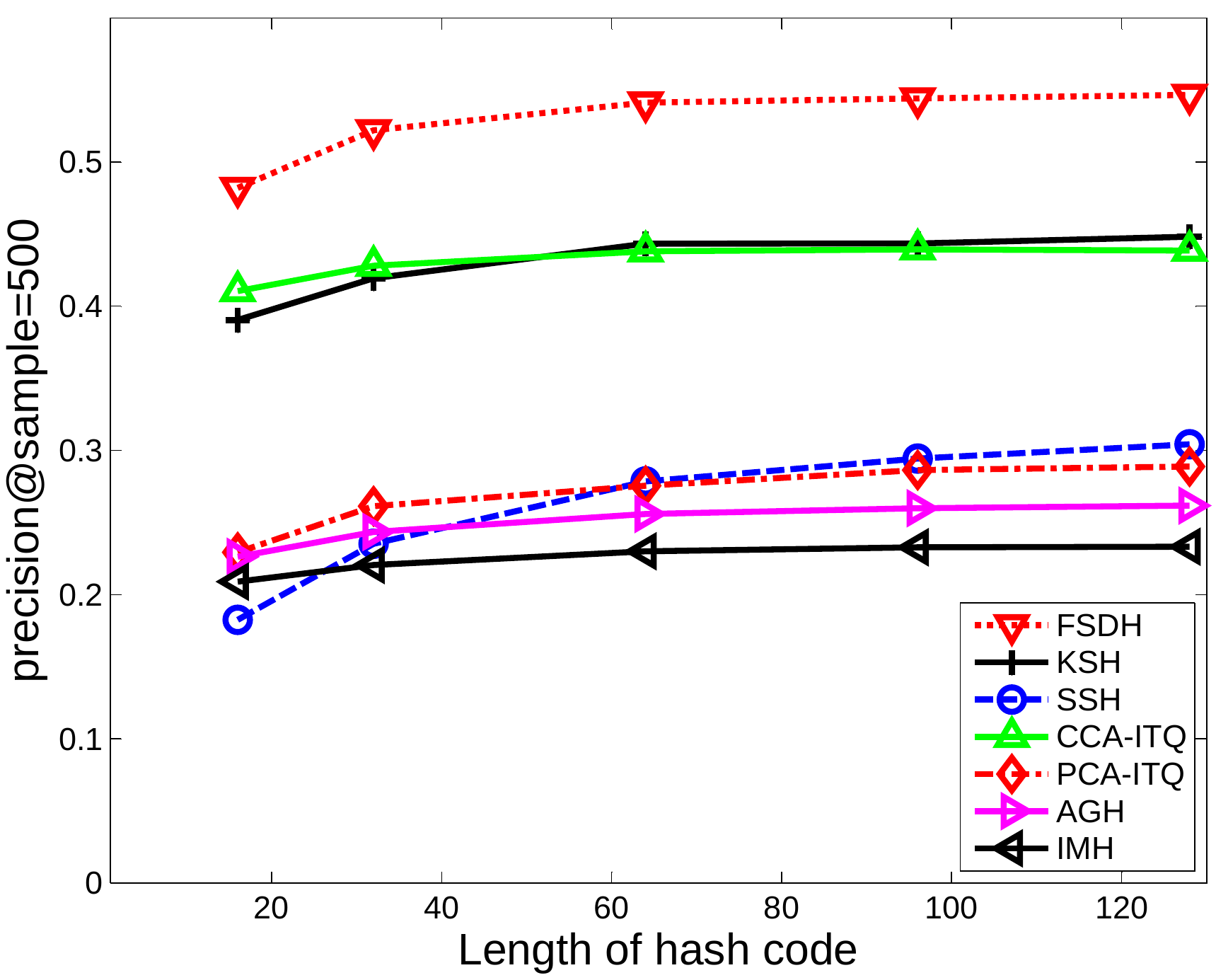}}
\end{center}
   \caption{Precision@sample=500 as functions of the number of hashing bits (16, 32, 64, 96, 128) on the CIFAR-10 database.}
\label{fig:2}
\end{figure}

\begin{figure}
\begin{center}
%%\fbox{\rule{0pt}{2in} \rule{.9\linewidth}{0pt}}
%\includegraphics[scale=0.4]{Fig1.pdf}
\scalebox{0.36}{\includegraphics{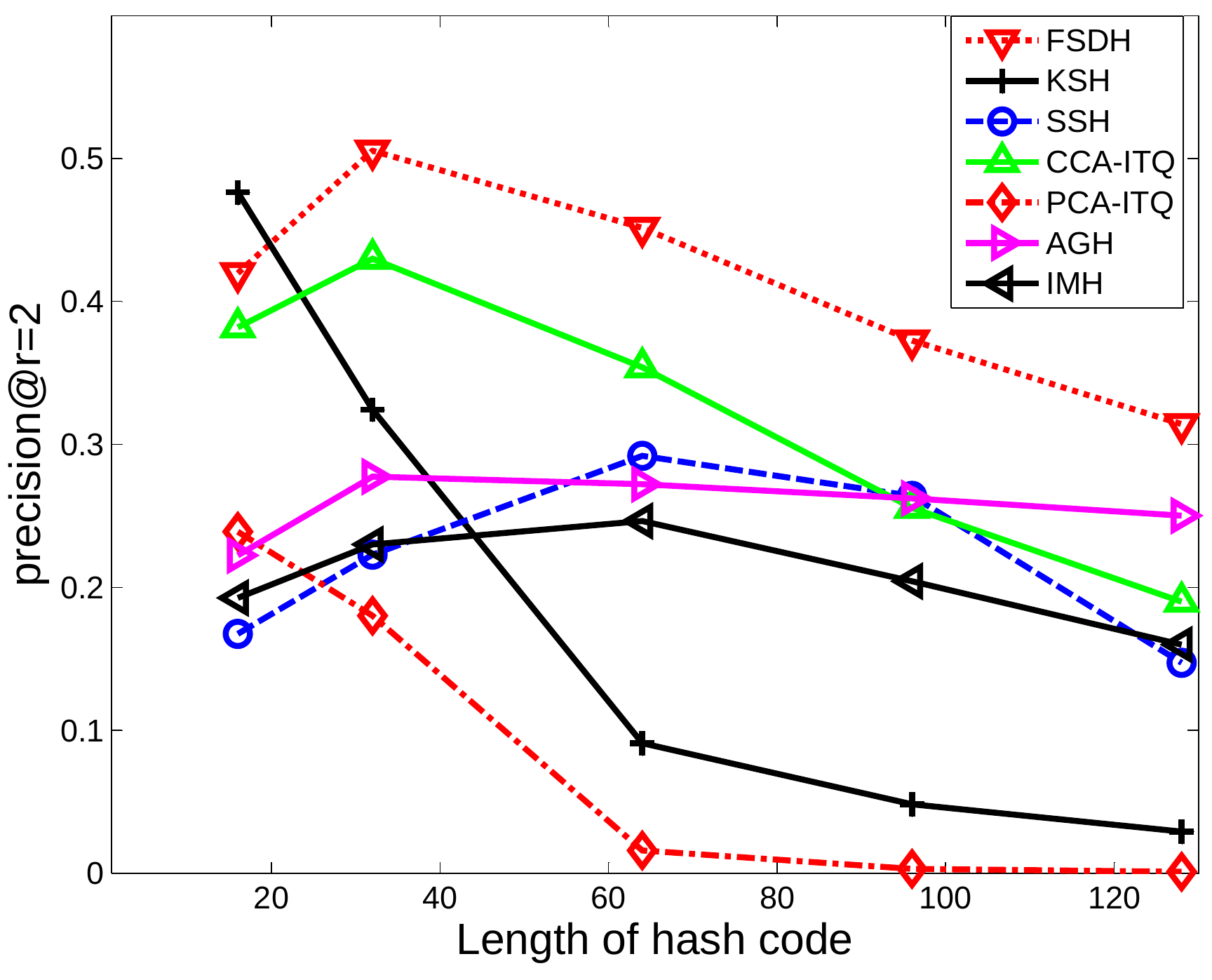}}
\end{center}
   \caption{Precision of Hamming radius 2 as functions of the number of hashing bits (16, 32, 64, 96, 128) on the CIFAR-10 database.}
\label{fig:3}
\end{figure}

\begin{figure}
\begin{center}
%%\fbox{\rule{0pt}{2in} \rule{.9\linewidth}{0pt}}
%\includegraphics[scale=0.4]{Fig1.pdf}
\scalebox{0.36}{\includegraphics{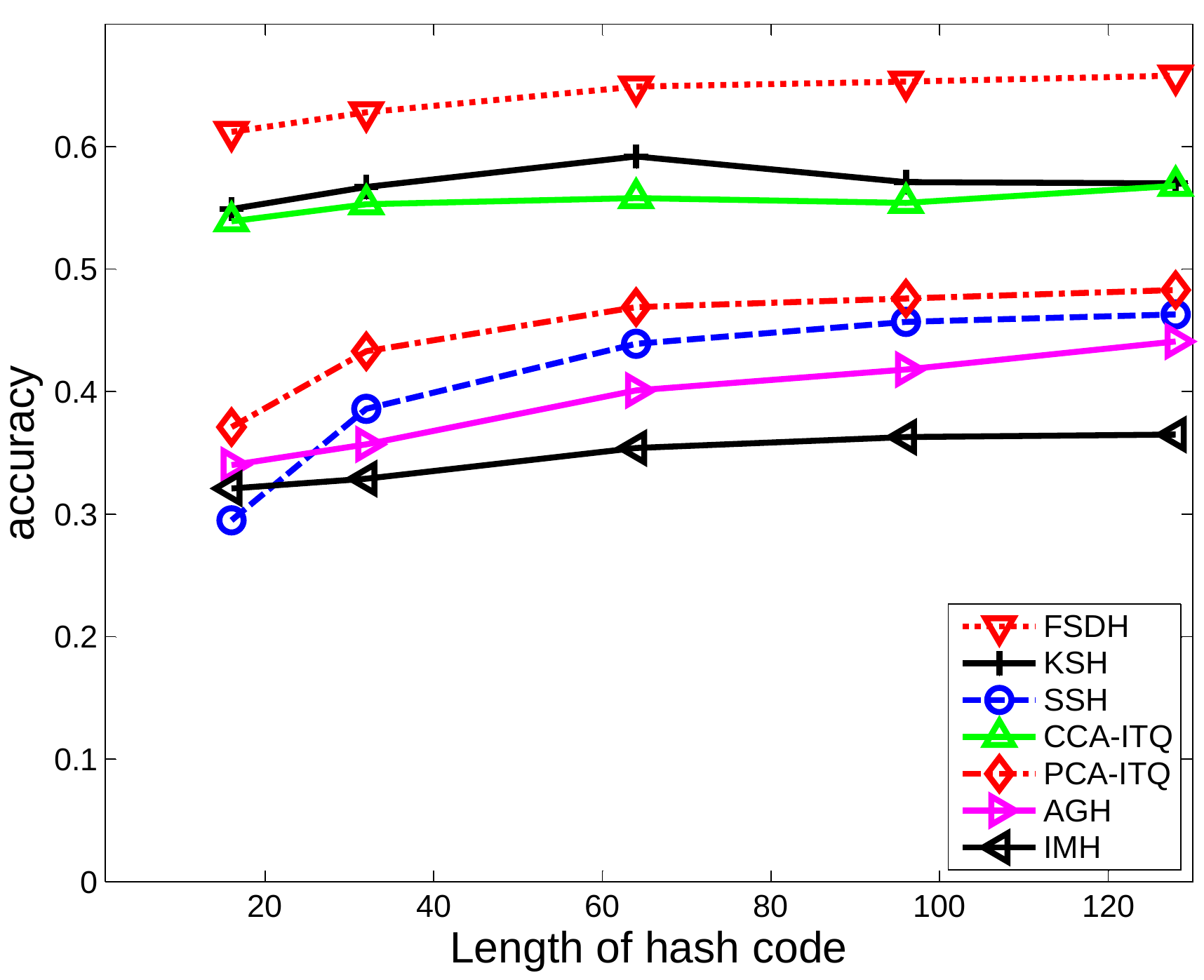}}
\end{center}
   \caption{Accuracy as functions of the number of hashing bits (16, 32, 64, 96, 128) on the CIFAR-10 database.}
\label{fig:4}
\end{figure}

As a subset of the well-known 80M tiny image collection \cite{torralba200880}, CIFAR-10 contains 60,000 images from 10 classes with 6,000 instances for each class. Each image is represented by a 512-dimensional GIST feature vector \cite{oliva2001modeling}. The entire dataset is split into a test set with 1,000 examples and a training set with all remaining examples.

The experimental results on CIFAR-10 are shown in Table \ref{tab:1} when the number of hashing bits is 128. Precision, recall, F-measure of Hamming distance within radius 2, MAP, accuracy, training time, and test time are presented. For SSH, we utilize 5,000 labeled instances for similarity matrix construction. FSDH outperforms SDH in terms of precision, recall, F-measure, and accuracy. FSDH takes only about half a minute to train on all 59,000 training examples. In contrast, KSH and FastHash take about 3 hours and 20 minutes, respectively. CCA-ITQ, SSH, PCA-ITQ, AGH, and IMH are also very efficient; however, their performance is generally worse than FSDH. The precision at 500 examples, precision of Hamming radius 2, and accuracy versus the number of hashing bits are shown in Figs. \ref{fig:2}-\ref{fig:4}, respectively (only some comparison methods are shown due to space limitations). With respect to precision of Hamming radius 2, FSDH outperforms the other methods when the number of hashing bits is larger than 32, and KSH performs the best when the number of hashing bits is 16. FSDH outperforms the other methods in terms of accuracy and precision at 500 examples, highlighting the effectiveness of our method.

A critical advantage of FSDH is that it is very fast. For example, FSDH and SDH take 32.8 and 406.4 seconds, respectively, when the number of hashing bits is 128. Thus, FSDH is about 12-times faster than SDH in this case. FastHash also performs very well; however, it is much slower than FSDH. For example, FastHash takes 1183.3 seconds with 128 hashing bits. Thus, FSDH is about 36-times faster than FastHash in this case.

\subsection{Experimental results on MNIST}
\begin{figure}
\begin{center}
%%\fbox{\rule{0pt}{2in} \rule{.9\linewidth}{0pt}}
%\includegraphics[scale=0.4]{Fig1.pdf}
\scalebox{0.36}{\includegraphics{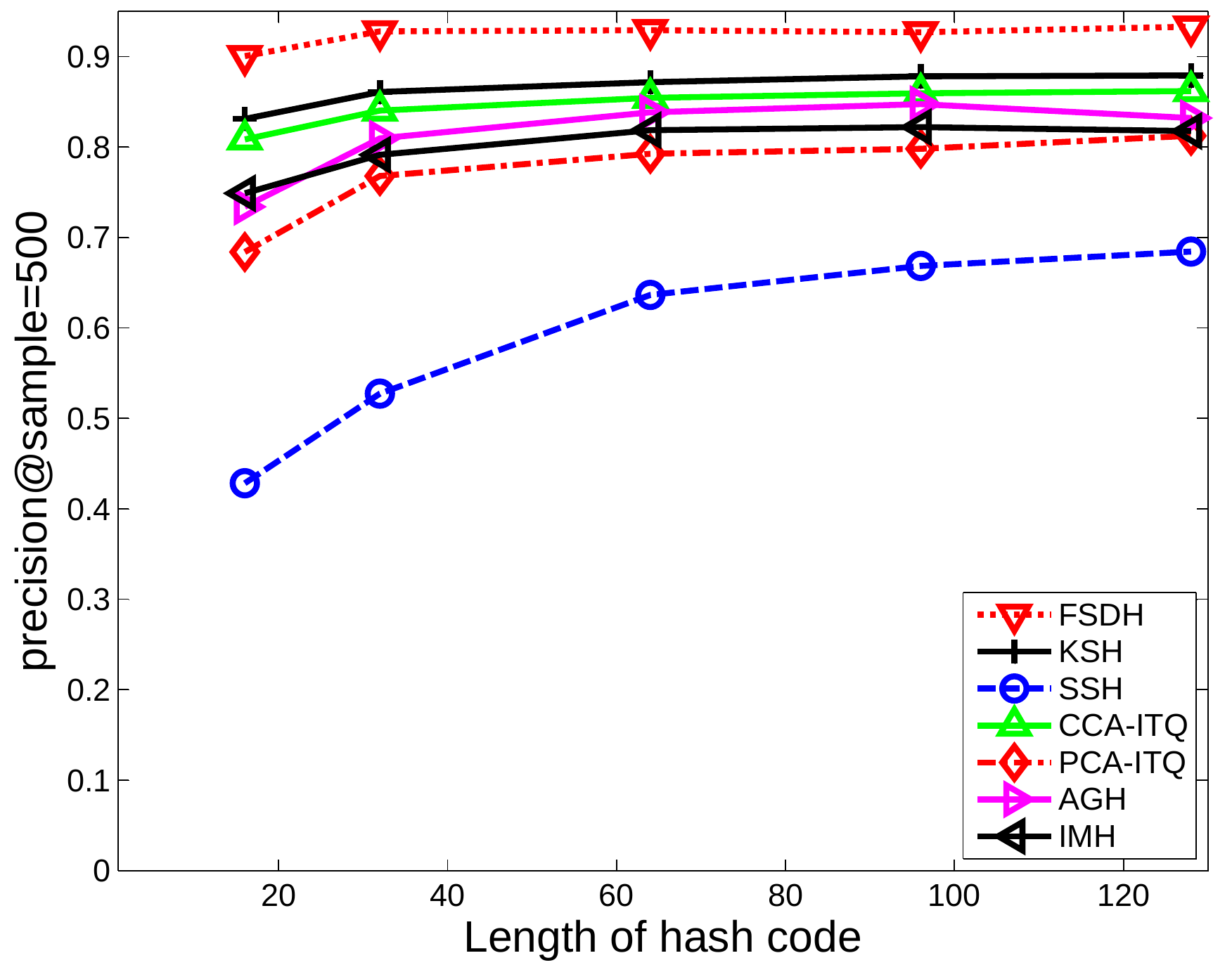}}
\end{center}
   \caption{Precision@sample=500 as functions of the number of hashing bits (16, 32, 64, 96, 128) on the MNIST database.}
\label{fig:6}
\end{figure}

\begin{figure}
\begin{center}
%%\fbox{\rule{0pt}{2in} \rule{.9\linewidth}{0pt}}
%\includegraphics[scale=0.4]{Fig1.pdf}
\scalebox{0.36}{\includegraphics{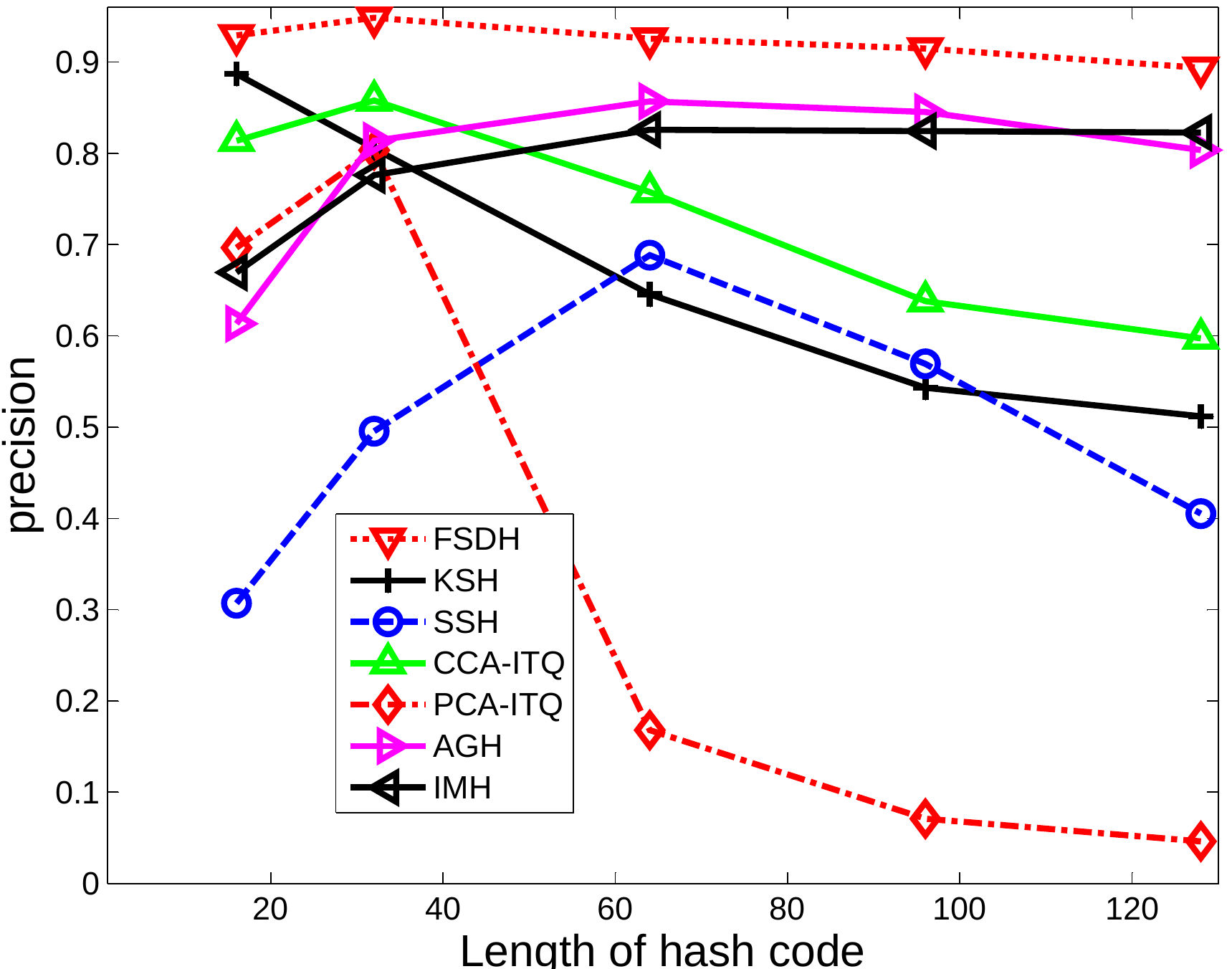}}
\end{center}
   \caption{Precision of Hamming radius 2 as functions of the number of hashing bits (16, 32, 64, 96, 128) on the MNIST database.}
\label{fig:7}
\end{figure}

\begin{figure}
\begin{center}
%%\fbox{\rule{0pt}{2in} \rule{.9\linewidth}{0pt}}
%\includegraphics[scale=0.4]{Fig1.pdf}
\scalebox{0.36}{\includegraphics{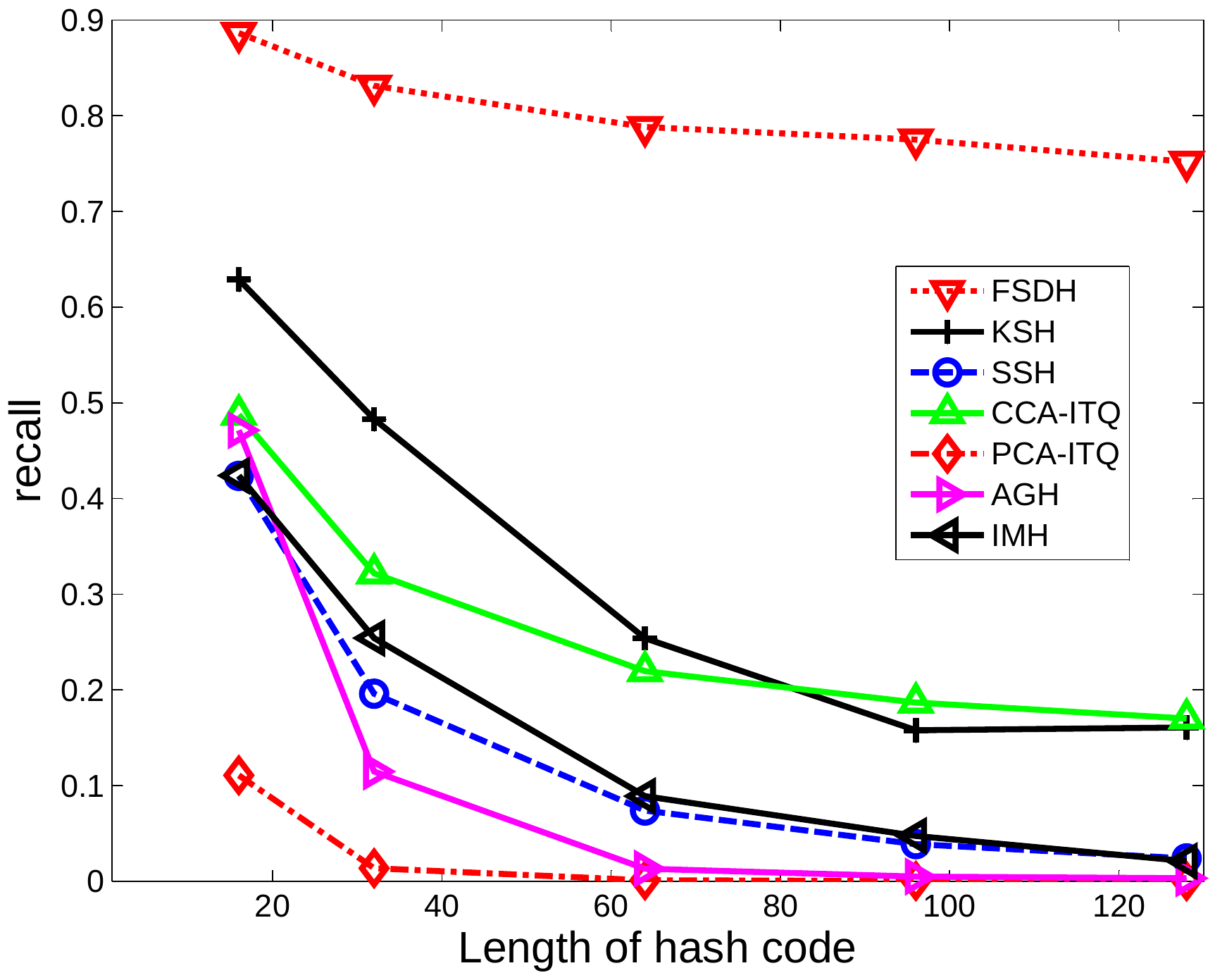}}
\end{center}
   \caption{Recall of Hamming radius 2 as functions of the number of hashing bits (16, 32, 64, 96, 128) on the MNIST database.}
\label{fig:8}
\end{figure}

\begin{figure}
\begin{center}
%%\fbox{\rule{0pt}{2in} \rule{.9\linewidth}{0pt}}
%\includegraphics[scale=0.4]{Fig1.pdf}
\scalebox{0.36}{\includegraphics{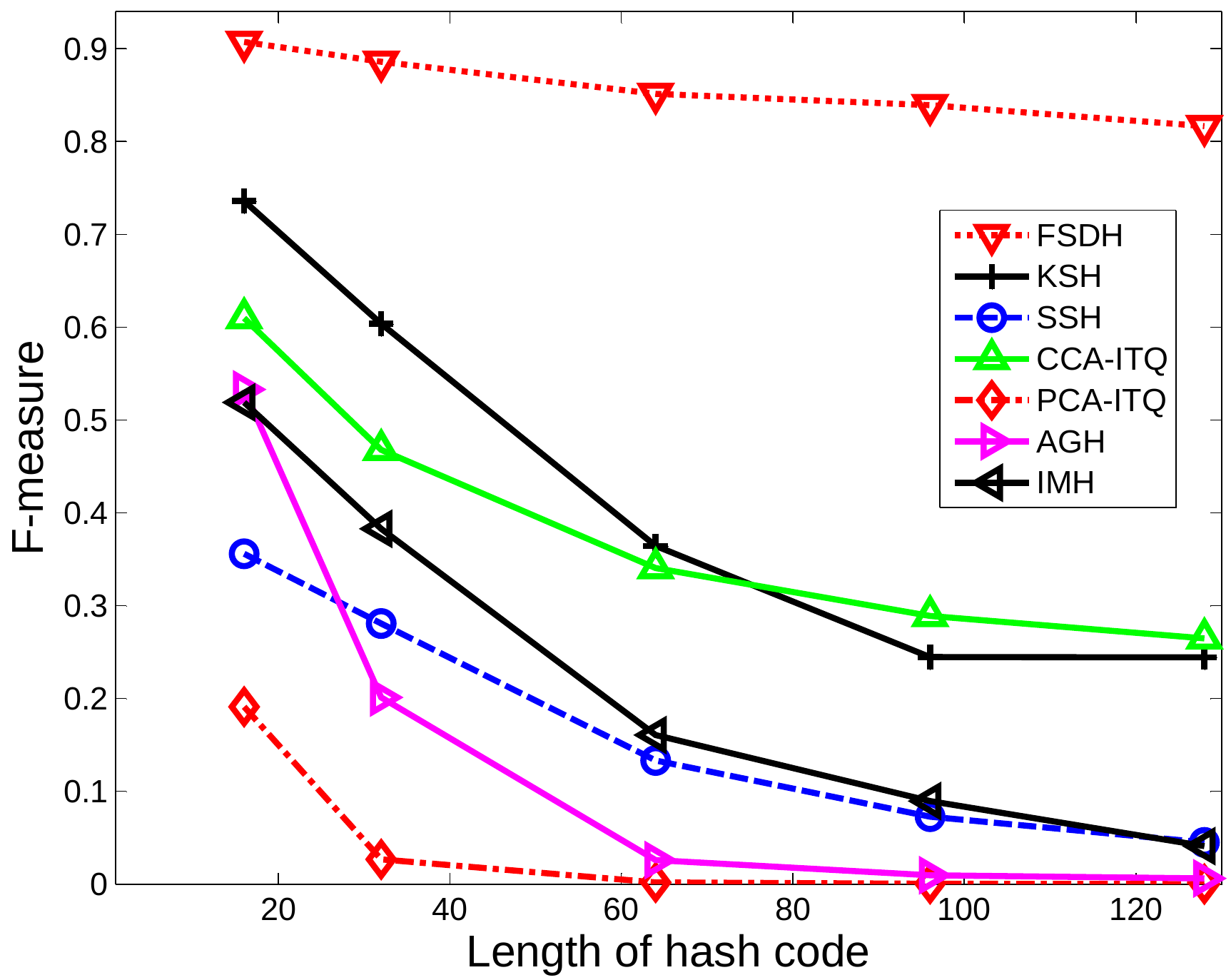}}
\end{center}
   \caption{F-measure of Hamming radius 2 as functions of the number of hashing bits (16, 32, 64, 96, 128) on the MNIST database.}
\label{fig:9}
\end{figure}

\begin{figure}
\begin{center}
%%\fbox{\rule{0pt}{2in} \rule{.9\linewidth}{0pt}}
%\includegraphics[scale=0.4]{Fig1.pdf}
\scalebox{0.36}{\includegraphics{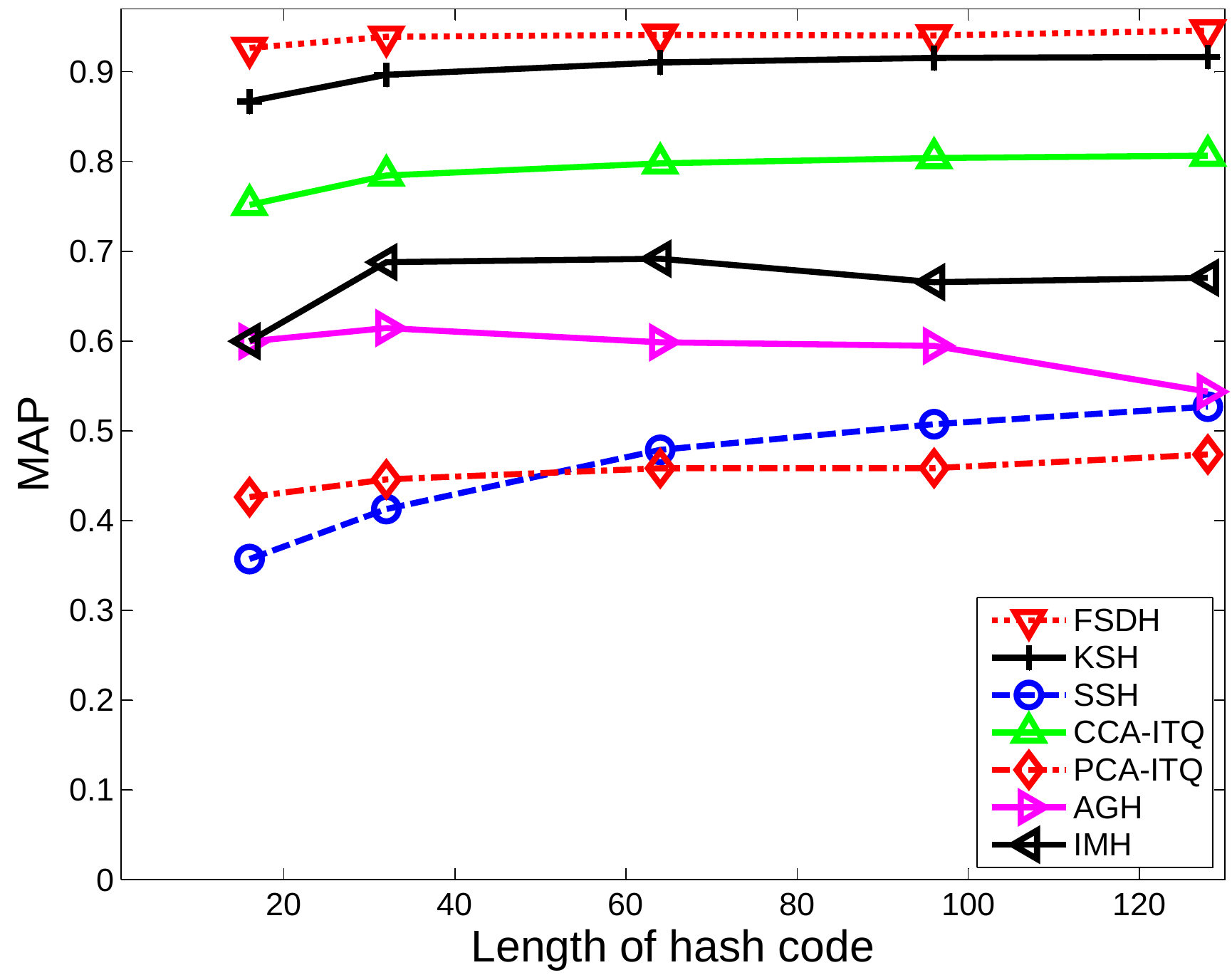}}
\end{center}
   \caption{MAP as functions of the number of hashing bits (16, 32, 64, 96, 128) on the MNIST database.}
\label{fig:10}
\end{figure}

\begin{figure}
\begin{center}
%%\fbox{\rule{0pt}{2in} \rule{.9\linewidth}{0pt}}
%\includegraphics[scale=0.4]{Fig1.pdf}
\scalebox{0.36}{\includegraphics{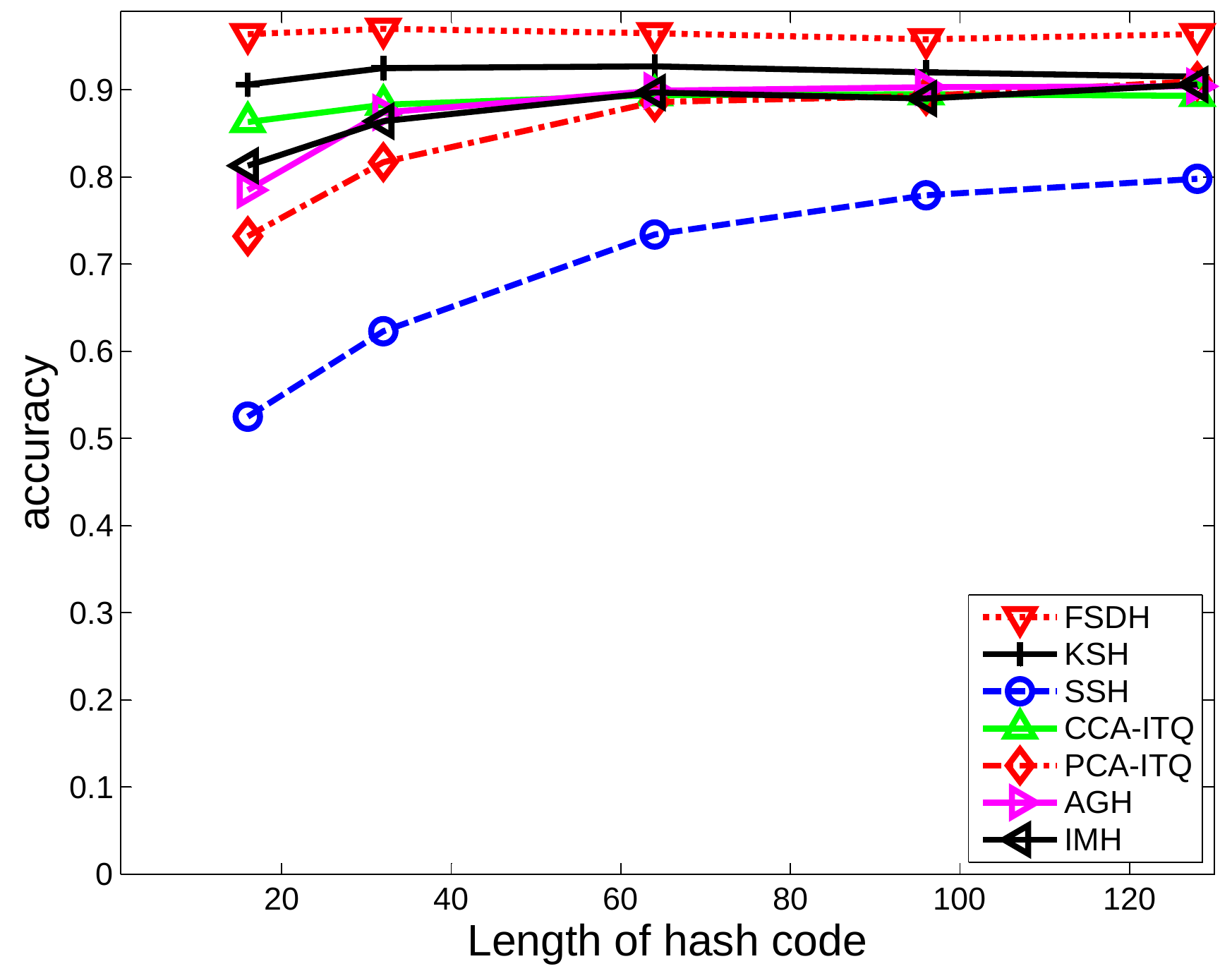}}
\end{center}
   \caption{Accuracy as functions of the number of hashing bits (16, 32, 64, 96, 128) on the MNIST database.}
\label{fig:11}
\end{figure}

\begin{table*}
\begin{center}
\scalebox{1}[1]{
\begin{tabular}{|l|c|c|c|c|c|c|c|}
\hline
Method & precision@$r$=2 & recall@$r$=2 & F-measure@$r$=2 & MAP & accuracy & training time & test time\\
\hline\hline
FSDH & 0.9256 & \textbf{0.7881} & \textbf{0.8513} & 0.9410 & 0.965 & 30.7 & 4.3e-6\\
SDH  & \textbf{0.9269} & 0.7711 & 0.8419 & 0.9397 & 0.963 & 128 & 5.1e-6\\
BRE  & 0.3850 & 0.0011 & 0.0021 & 0.4211 & 0.839 & 24060.6 & 9.3e-5 \\
KSH  & 0.6454 & 0.2539 & 0.3644 & 0.9103 & 0.927 & 1324.7 & 7.6e-5\\
SSH  & 0.6883 & 0.0738 & 0.1332 & 0.4787 & 0.734 & 260.2 & 5.7e-6\\
CCA-ITQ  & 0.7575 & 0.2196 & 0.3405 & 0.7978 & 0.894 & 16.6 & \textbf{4.1e-7}\\
FastHash  &0.8680  & 0.6735 & 0.7585 &\textbf{0.9813}  & \textbf{0.972} & 4661.1 &0.0012 \\
PCA-ITQ  & 0.1680 & 9.3e-4 & 0.0018 & 0.4581 & 0.886 & 10.1 & 4.5e-7\\
AGH  & 0.8568 & 0.0131 & 0.0258 & 0.5984 & 0.899 & \textbf{6.9} & 6.4e-5\\
IMH  & 0.8258 & 0.0889 & 0.1606 & 0.6916 & 0.897 & 32.2 & 6.4e-5\\
\hline
\end{tabular}}
\end{center}
\caption{Experimental results on the MNIST database when the number of hashing bits is 64. Training and test times are in seconds. The best results are highlighted in bold face.}\label{tab:2}
\end{table*}
MNIST contains 70,000 784-dimensional handwritten digit images from ¡®0¡¯ to ¡®9¡¯. Each image is cropped and normalized to 28$\times$28. The dataset is split into a training set with 69,000 examples and a test set with all remaining examples. The experimental results on MNIST are shown in Table \ref{tab:2}. FSDH performs best in terms of recall and F-measure, SDH performs the best in terms of precision, and FastHash performs the best in terms of MAP and accuracy. However, FastHash is much slower than FSDH, taking 4661.1 and 30.7 seconds to train, respectively. Thus, FSDH is about 151-times faster than FastHash in this setting. The precision@sample = 500, precision of Hamming radius 2, recall of Hamming radius 2, F-measure of Hamming radius 2, MAP, and accuracy curves are shown in Figs. \ref{fig:6}-\ref{fig:11}, respectively (only some methods are shown due to space limitations). FSDH outperforms all other methods.

\subsection{Experimental results on the FRGC face database }
\begin{figure}
\begin{center}
%%\fbox{\rule{0pt}{2in} \rule{.9\linewidth}{0pt}}
%\includegraphics[scale=0.4]{Fig1.pdf}
\scalebox{0.45}{\includegraphics{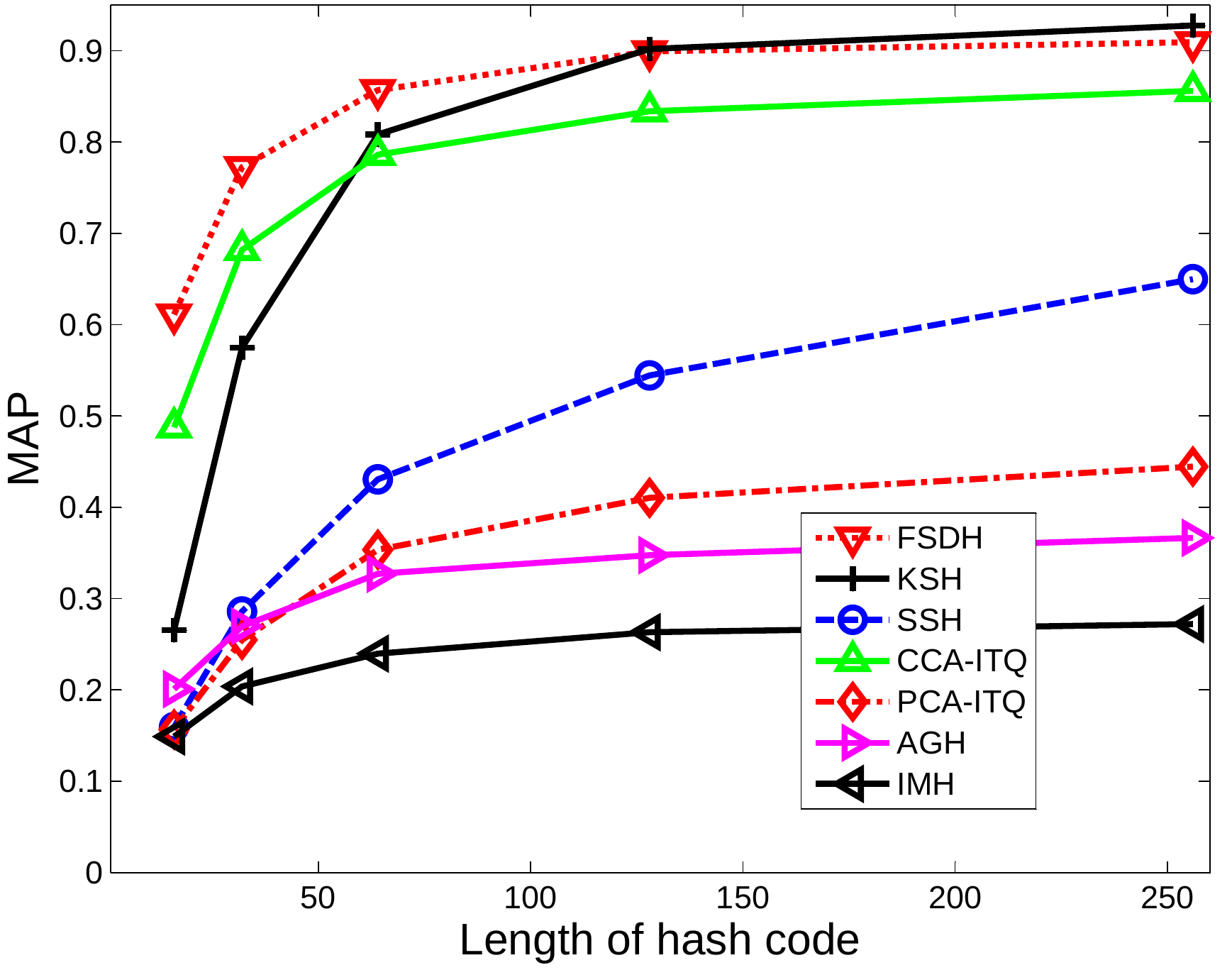}}
\end{center}
   \caption{MAP as functions of the number of hashing bits (16, 32, 64, 128, 256) on the FRGC database.}
\label{fig:13}
\end{figure}

\begin{figure}
\begin{center}
%%\fbox{\rule{0pt}{2in} \rule{.9\linewidth}{0pt}}
%\includegraphics[scale=0.4]{Fig1.pdf}
\scalebox{0.45}{\includegraphics{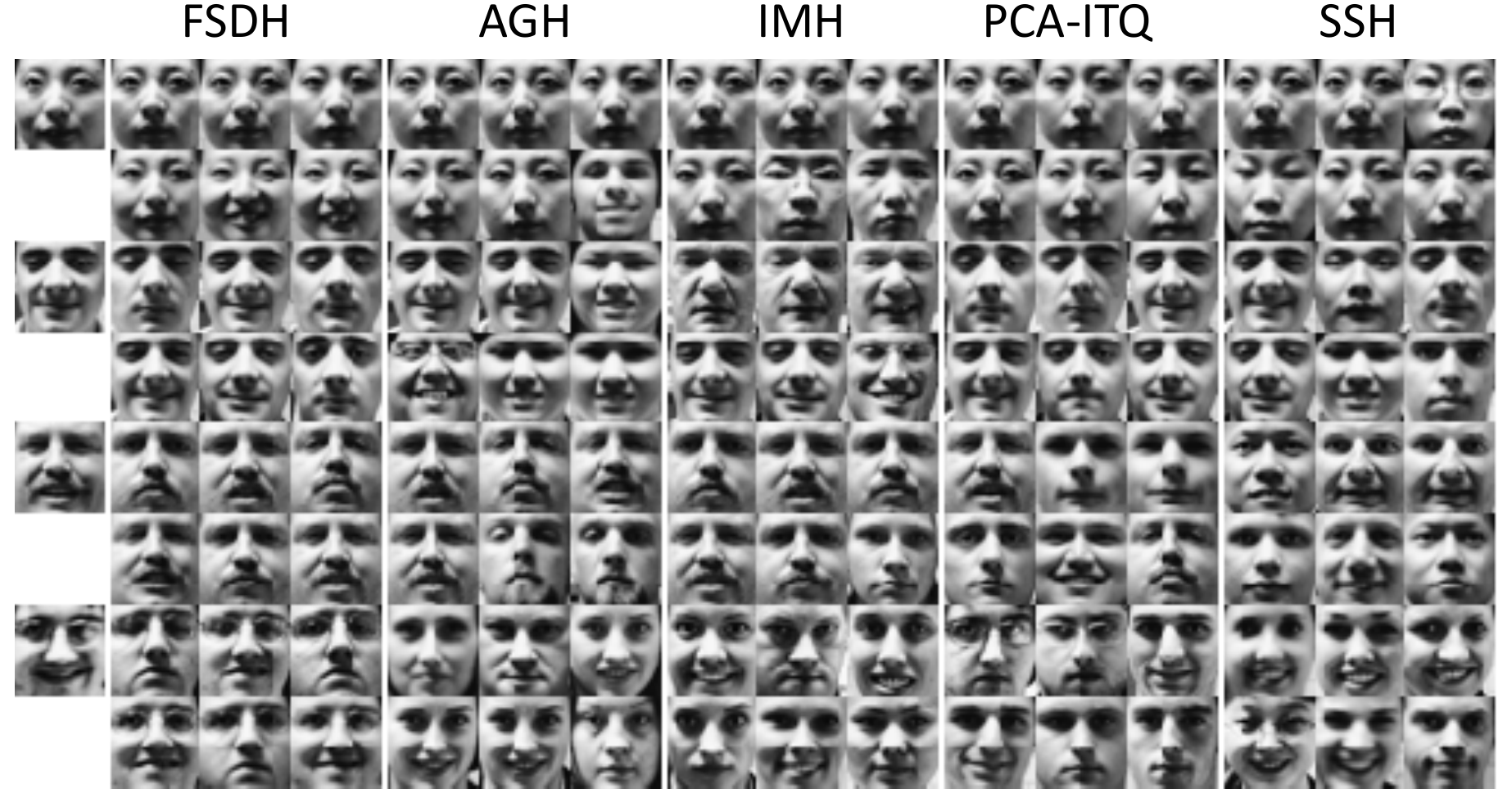}}
\end{center}
   \caption{Top retrieved 6 images of 4 queries returned by various hashing methods on the FRGC data base. The image on the first column is the query instance. From left to right are the retrieved images by FSDH, AGH, IMH, PCA-ITQ and SSH when 16-bit binary codes are utilized for search. }
\label{fig:14}
\end{figure}

\begin{table*}
\begin{center}
\scalebox{1}[1]{
\begin{tabular}{|l|c|c|c|c|c|c|c|}
\hline
Method & precision@$r$=2 & recall@$r$=2 & F-measure@$r$=2 & MAP & accuracy & training time & test time\\
\hline\hline
FSDH & \textbf{0.4400} & \textbf{0.3810} & \textbf{0.4084} & 0.7725 & 0.753 & 1.2 & 2.1e-6\\
SDH  & \textbf{0.4400} & 0.3740 & 0.4044 & \textbf{0.7775} & 0.743 & 2.0 & 2.2e-6\\
BRE  & 0.1690 & 0.0881 & 0.1158 & 0.1458 & 0.252 & 216.7 & 1.8e-5\\
KSH  & 0.2383 & 0.0752 & 0.1144 & 0.5748 & 0.642 & 736.2 & 1.2e-4\\
SSH  & 0.1856 & 0.0464 & 0.0743 & 0.2858 & 0.45 & 9.3 & 7.4e-6\\
CCA-ITQ  & 0.3333 & 0.1202 & 0.1767 & 0.6819 & \textbf{0.783} & 1.0 & \textbf{1.7e-7}\\
FastHash  & 0.0400 & 0.0057 & 0.0100 & 0.2523 & 0.51 & 133.6 & 1.4e-3\\
PCA-ITQ  & 0.2172 & 0.0726 & 0.1088 & 0.2548 & 0.452 & \textbf{0.3} & 2.2e-7\\
AGH  & 0.2366 & 0.3398 & 0.2789 & 0.2696 & 0.391 & 2.6 & 1.6e-4\\
IMH  & 0.1447 & 0.3143 & 0.1982 & 0.2038 & 0.287 & 36.8 & 9.1e-5\\
\hline
\end{tabular}}
\end{center}
\caption{Experimental results on the FRGC face database when the number of hashing bits is 32. Training and test times are in seconds. The best results are highlighted in bold face.}\label{tab:3}
\end{table*}

The FRGC version two face database \cite{phillips2005overview} is a challenging and large-scale benchmark face database with 8014 face images from 466 individuals in the query set for FRGC experiment 4. These uncontrolled images demonstrate variations in blurring, illumination, expression, and time. In our experiment, only individuals represented by over 10 images in the database are used (3160 images from 316 individuals). Each image is cropped and resized to 32$\times$32 pixels by fixing the eye positions in all experiments, with 256 gray levels per pixel. For each person, seven images are randomly selected for training and the remainder used for testing.

Experimental results on FRGC are shown in Table \ref{tab:3}. FSDH performs the best in terms of precision, recall, and F-measure, while SDH performs best for MAP and precision. CCA-ITQ performs the best in terms of accuracy. The MAP versus the number of hashing bits is presented in Fig. \ref{fig:13}; due to space limitations, only representative methods are shown. FSDH performs the best with 64 hashing bits, while KSH outperforms the other methods when the number of hashing bits is greater than or equals 128. Furthermore, the MAP of all methods increases as the number of hashing bits increases, perhaps due to more information being encoded in the hash code as the number of hashing bits increases. Thus, the face image is represented by the hash code in a more discriminative and informative way. Several sample query images and the retrieved neighbors when 16 bits are utilized to learn binary codes for various hashing algorithms are shown in Fig. \ref{fig:14}. FSDH delivers better search results since higher semantic relevance is obtained in the top retrieved instances.

\section{Conclusion}
In this paper, we present a new data-dependent hashing algorithm called ``fast supervised discrete hashing" (FSDH) based on ``supervised discrete hashing" (SDH). FSDH regresses the class label to the corresponding hash code, which not only makes it faster but also improves overall performance compared to SDH. Experimental results on image classification and face recognition datasets show that FSDH is very efficient and effective.

As an effective and efficient nonlinear feature extraction algorithm, this method can also be applied to other practical applications, especially those involving large-scale data, for example, large-scale mobile video retrieval and visual tracking. Another interesting application of FSDH would be compressing the high-dimensional features into short binary codes, which could significantly speed up large-scale visual tasks such as ImageNet image classification.

In the supervised discrete hashing framework, the hash code is approximated via nonlinear embedding. Deep learning is a current ``hot topic", and using deep learning as a nonlinear embedding tool in the SDH framework would be intuitive. However, embedding deep learning in the SDH framework as the nonlinear embedding technique slows the original method. We are now investigating how to effectively and efficiently combine them.
% or


\ifCLASSOPTIONcaptionsoff
  \newpage
\fi



%\bibliographystyle{IEEEtran}
\bibliographystyle{ieeetr}
\bibliography{a}

%\end{IEEEbiography}

%\newpage


\end{document}